\documentclass[acmlarge,nonacm]{acmart}
 \usepackage{hyperref}
	\usepackage{color,soul}
 \usepackage{graphicx}
	\usepackage{diagbox}
	\usepackage{makecell}
\usepackage[figuresright]{rotating}
\usepackage{lscape}
	\usepackage{multirow,multicol}
	\usepackage{textcomp}
 \usepackage{setspace}
	\usepackage{enumitem}
 \usepackage{outlines}
	\usepackage{multicol}
    \usepackage{multirow}
    \usepackage{tabularx}
    \usepackage{colortbl}
    \usepackage{caption}
    \usepackage{subfigure}
    \usepackage{lipsum}
    \usepackage[utf8]{inputenc}
    \usepackage{algorithm2e}
    \usepackage{array}
    \usepackage{xcolor}
    \usepackage[flushleft]{threeparttable}

\usepackage{longtable}
\usepackage{geometry}
\usepackage{pdflscape}
\usepackage{booktabs} 

\renewcommand{\hl}[1]{{#1}}

\newcommand{\PreserveBackslash}[1]{\let\temp=\\#1\let\\=\temp}
\newcolumntype{C}[1]{>{\PreserveBackslash\centering}p{#1}}

\AtBeginDocument{%
  \providecommand\BibTeX{{%
    \normalfont B\kern-0.5em{\scshape i\kern-0.25em b}\kern-0.8em\TeX}}}

\begin{document}

\title{PALLM: Evaluating and Enhancing \underline{PALL}iative Care Conversations with \underline{L}arge \underline{L}anguage \underline{M}odels}


\author{\href{https://orcid.org/0000-0002-1611-2053}{Zhiyuan Wang}}
\authornote{Both authors contributed equally to this research.}
\authornote{Corresponding author: Zhiyuan Wang (vmf9pr@virginia.edu)}
\email{vmf9pr@virginia.edu}
\affiliation{%
  \institution{Department of Systems and Information Engineering, University of Virginia}
  \streetaddress{151 Engineer's Way}
  \city{Charlottesville}
  \state{Virginia}
  \country{USA}
  \postcode{22904}
}

\author{Fangxu Yuan}
\authornotemark[1]
\email{phh3pb@virginia.edu}
\affiliation{%
  \institution{Department of Systems and Information Engineering, University of Virginia}
  \streetaddress{151 Engineer's Way}
  \city{Charlottesville}
  \state{Virginia}
  \country{USA}
  \postcode{22903}
}

\author{Virginia LeBaron}
\email{vtl6k@virginia.edu}
\affiliation{%
  \institution{School of Nursing, University of Virginia}
  \streetaddress{225 Jeanette Lancaster Way}
  \city{Charlottesville}
  \state{Virginia}
  \country{USA}
  \postcode{22903}
}

\author{Tabor Flickinger}
\email{tes3j@virginia.edu}
\affiliation{%
  \institution{School of Medicine, University of Virginia}
  \streetaddress{1340 Jefferson Park Ave}
  \city{Charlottesville}
  \state{Virginia}
  \country{USA}
  \postcode{22903}
}

\author{\href{https://orcid.org/0000-0001-8224-5164}{Laura E. Barnes}}
\email{lb3dp@virginia.edu}
\affiliation{%
  \institution{Department of Systems and Information Engineering, University of Virginia}
  \streetaddress{151 Engineer's Way}
  \city{Charlottesville}
  \state{Virginia}
  \country{USA}
  \postcode{22904}
}

\renewcommand{\shortauthors}{Zhiyuan Wang et al.}

\begin{teaserfigure}
    \centering
    \includegraphics[width=0.9\linewidth]{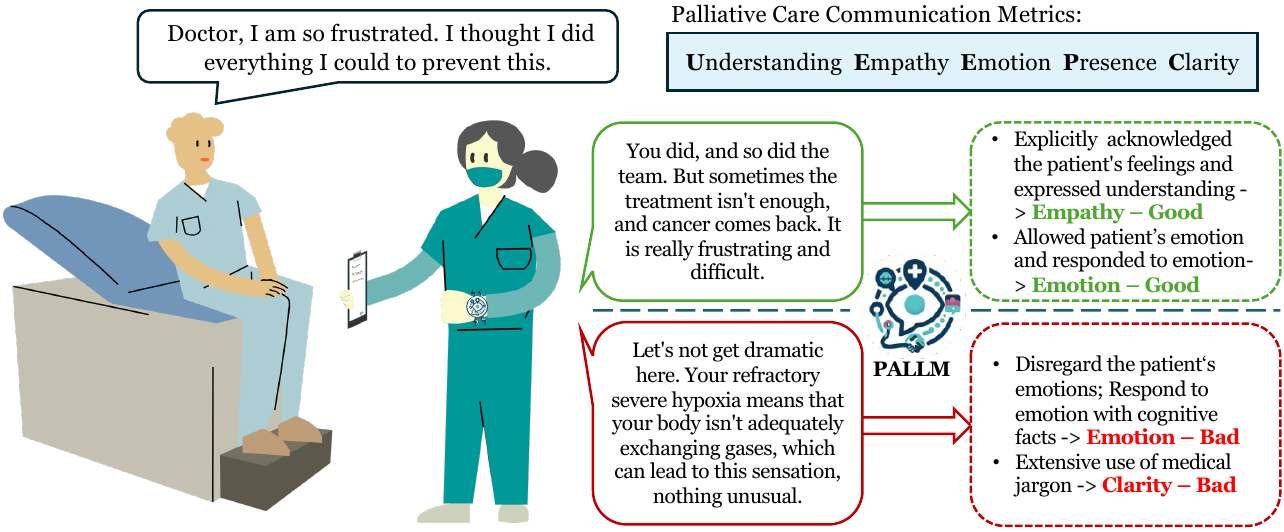}
    \caption{Illustration of LLM-empowered evaluation of palliative care communication metrics.}
    \label{fig:intro}
\end{teaserfigure}
\begin{abstract}
Effective patient-provider communication is crucial in clinical care, directly impacting patient outcomes and quality of life. Traditional evaluation methods, \hl{such as human ratings, patient feedback, and provider self-assessments, are often limited by high costs and scalability issues. Although existing natural language processing (NLP) techniques show promise, they struggle with the nuances of clinical communication and require sensitive clinical data for training, reducing their effectiveness in real-world applications.} Emerging large language models (LLMs) offer a new approach to assessing complex communication metrics, with the potential to advance the field through integration into passive sensing and just-in-time intervention systems. This study explores LLMs as evaluators of palliative care communication quality, leveraging their linguistic, in-context learning, and reasoning capabilities. \hl{Specifically, using simulated scripts crafted and labeled by healthcare professionals, we test proprietary models (e.g., GPT-4) and fine-tune open-source LLMs (e.g., LLaMA2) with a synthetic dataset generated by GPT-4 to evaluate clinical conversations, to identify key metrics such as `understanding' and `empathy'. Our findings demonstrated LLMs' superior performance in evaluating clinical communication, providing actionable feedback with reasoning, and demonstrating the feasibility and practical viability of developing in-house LLMs.} This research highlights LLMs' potential to enhance patient-provider interactions and lays the groundwork for downstream steps in \hl{developing LLM-empowered clinical health systems}.
\end{abstract}


\maketitle
\section{Introduction}

Clinical conversations between patients and healthcare providers are important to care, acting as the primary means for exchanging health information, making informed treatment decisions, and offering emotional support \cite{parker2000improving,curtis2013effect}. These interactions are especially important in palliative care settings \cite{morrison2004palliative}, where patients face serious illnesses like cancer and endure symptoms like cancer pain.  Conversations go beyond basic medical information, incorporating essential emotional, social, and spiritual support that significantly impacts their health and quality of life \cite{kavalieratos2016association,lebaron2024caring}. In this context, effective communication is crucial for managing symptoms, making treatment decisions, and providing emotional support, leading to increased patient satisfaction, less distress, and better quality of life \cite{halpern2003clinical,geerse2019qualitative}. It necessitates a blend of empathy, active listening, and the sensitive navigation of topics such as prognosis and end-of-life preferences, underscoring the need for healthcare providers to cultivate these skills through targeted training \cite{goldsmith2013palliative}. Given the substantial impact on patient and family experiences, the pursuit of quality in patient-provider communication emerges as a paramount goal, highlighting the need for actionable evaluation and feedback \cite{hojat2016empathy}.

Traditional methods for training and assessing healthcare providers' communication skills, which highly rely on human raters for observation and feedback \cite{tulsky2017research,brighton2017systematic}, are beneficial but struggle with scalability and cost, hard to expand and expensive for healthcare organizations \cite{sanders2020quality,byerly1969nurse,sennekamp2012development,meinert2021exploring}. Recent studies \cite{lebaron2022exploring,lebaron2023commsense,flickinger2023evidence,wang2024commsense} have investigated using natural language processing (NLP) models (like open-ended question detection \cite{wan2016modeling}, emotion detection \cite{hosseini2021takes}, and semantic feature extraction tools \cite{tausczik2010psychological}) to evaluate communication skills by comparing linguistic features with specific rules. However, these NLP approaches require sensitive clinical data for training and have limited capacity to fully capture the granular and complex nature of clinical communication \cite{wang2024commsense}, limiting their effectiveness and practicality.

Recent advances in generative artificial intelligence (GenAI), particularly large language models (LLMs) \cite{brown2020language}, have shown great promise in various natural language processing tasks like text generation \cite{brown2020language,wang2024grammar,yu2024large}, summarization \cite{tang2023evaluating,zhang2024benchmarking,zhang2024benchmarking}, reasoning \cite{wei2022chain,wang2022self,yao2024tree}, and embedding specialized knowledge and factuality \cite{lee2022factuality,wang2023survey,ding2023parameter}, including in medical and healthcare fields \cite{thirunavukarasu2023large,li2024chatgpt}. These models, trained on extensive data, can interpret and respond to complex in-context prompts, supporting multitasking  in healthcare communication. LLMs can be securely and remotely hosted and integrated into devices like smartwatches \cite{li2024personal}, \hl{offering a scalable and efficient method for analyzing and improving clinical communication. As shown in Figure \ref{fig:intro}, we envision this setup provides varying evaluations and feedback to healthcare providers, helping them improve interaction skills.}

However, several gaps limit LLMs’ application in evaluating clinical communication. Firstly, the capability of LLMs to grasp the interested subtleties of clinical communication, especially in palliative care, remains unexplored. Secondly, to the best of our knowledge, there is a notable absence of publicly available datasets containing clinical communication, hindering the development of task-specific language models. Finally, although proprietary LLMs like GPT-4 have shown exceptional effectiveness across various domains, the need to upload highly sensitive clinical data raises significant privacy and security concerns.

In this study, we focus on testing and tailoring LLMs for their ability to comprehend and evaluate clinical conversations. With eight simulated clinical communication scripts created and labelled by healthcare providers, we first utilize proprietary GPT-4 and different prompt engineering strategies (e.g., chain of thought) \cite{kojima2022large,wei2022chain,wang2022self} to assess clinical communication metrics including \textbf{understanding}, \textbf{empathy}, \textbf{emotion}, \textbf{presence}, and \textbf{clarity}, as outlined by systematically established criteria \cite{lebaron2023commsense,arnold2017oncotalk}. To enhance LLM utility in real-world clinics, we seek to fine-tune open-source Llama2-13b model, an open-source, smaller-scale LLM, for efficient \hl{in-house deployment}. Given the scarcity of real clinical conversation data, we generated a diverse synthetic dataset using GPT-4 for fine-tuning, with the goal of achieving performance comparable to the more complex GPT-series models. Our results indicate that GPT-series models achieve over 90\% balanced accuracy in identifying communication metrics. \hl{Additionally, the fine-tuned LLaMA2-13B models show strong performance, averaging 80\% accuracy and surpassing non-LLM NLP baselines, underscoring their potential for clinical communication analysis.}

Our contributions can be summarized as follows:

\begin{itemize}
    \item We \hl{explore and reveal} the feasibility of using LLMs to evaluate clinical conversations by integrating evaluation metrics and linguistic contexts into prompts and testing various prompting strategies.
    \item To address the scarcity of real-world clinical conversation datasets, we generated simulated data with healthcare professionals and a realistic synthetic dataset using GPT-4. \hl{We then fine-tuned open-source LLMs (Llama2-13b) on these datasets, achieving comparable performance. This demonstrates the feasibility and practical viability of developing in-house LLMs.}
    \item This paper establishes the groundwork for an LLM-powered system to evaluate clinical communication by understanding linguistic context, paving the way for future development of clinical assistant agents that support healthcare providers in improving communication skills.
\end{itemize}

\hl{We acknowledge that this foundational study, based on the analysis of eight human-generated scripts, may not provide a comprehensive benchmark for assessing LLMs in evaluating palliative care communication. Our team is currently investigating evaluations in real clinical settings and soliciting stakeholders' feedback on the ethical integration of these technologies into clinical practice. These efforts aim to unlock the potential of LLMs to improve patient-provider communication and contribute to better patient outcomes.}

\section{Background and Related Work}

This section reviews the background and related work on clinical communication and its evaluation, highlighting research gaps in digital health technology for this application and the potential of LLMs to advance the field.

\subsection{Clinical Communication Evaluation}

Traditionally, in real-world clinical practice, patient-provider communication can be evaluated based on providers' human raters, self-assessment, and/or patient's feedback.
\begin{itemize}
    \item \textbf{Human Raters:} Utilizing systems like the Roter Interaction Analysis System (RIAS) \cite{roter2002roter} and the Calgary-Cambridge guide \cite{kurtz2003marrying} to analyze structured communication. Although beneficial for educational purposes, the feasibility and scalability of these methods is constrained by the significant expenses associated with human labor and transcription, limiting their widespread use. Moreover, their validity is often compromised by inter-rater discrepancies in evaluations \cite{price2008assessing}.
    \item \textbf{Provider Self-Assessment}: As an alternative to human-rater assessment, providers' self-assessment used to evaluate clinical communication, which allows for reflection and self-improvement \cite{hawkins2012improving}. However, research has shown that professionals often have inaccurate perceptions of their skills, which lead to over- or underestimation of their abilities, hindering sustainable improvement \cite{davis2006accuracy}.
    \item \textbf{Patient Feedback}: Insights on the communication performance of providers can also be gathered through rating scales like the Consultation and Relational Empathy (CARE) Measure \cite{mercer2005relevance}. While valuable, this approach is subject to patients' subjective biases and influences by external, unrelated factors \cite{bikker2015measuring}, affecting the objective assessment.
\end{itemize}

Despite the value of these traditional evaluation methods, they often fall short in providing real-time, scalable, consistent assessments of clinical communication, largely due to their resource-intensive nature and subjective elements. The emergence of digital technologies, especially in natural language processing and ubiquitous computing, offers promising avenues to bridge these gaps, suggesting a shift towards more dynamic, scalable, and objective evaluation methods in clinical communication.

\subsection{Digital Technology in Clinical Communication and Evaluation}

Given the limitations of traditional human-feedback-based assessment methods, digital technology offers a scalable solution for improving the accessibility of provider-patient communication evaluation. In the realm of digital era, technological tools like clinical simulations \cite{foronda2014use,sweigart2014virtual}, virtual patients \cite{washburn2018teaching,shorey2020communication}, and Telehealth meetings \cite{richmond2017american,henry2018experienced} have been employed to enrich the education scenarios and enhance the providers' communication skills. However, still, the assessment of clinical communication often relies on human-based evaluations.

Amid these challenges, NLP technologies offered great potential for enhancing clinical communication and evaluation. NLP techniques have demonstrated their capacity in analyzing electronic health records \cite{yadav2018mining}, clinical interactions \cite{clarke2020things}, and treatment strategies \cite{yim2016natural}, providing insights into patient-provider interactions that were previously difficult to capture at scale along with advanced sensing techniques. Against this background, CommSense \cite{wang2024commsense,lebaron2023commsense} represents a pivotal project that identifies feasible communication metrics to be detected and analyzed by wearable sensing and NLP techniques to evaluate and improve palliative care conversations. Specifically, the five evaluation metrics systematically specified to rate the providers' communication skills include understanding, empathy, emotion, presence, and clarity, with detailed NLP operational rules for assessment listed in Table \ref{tab:criteria-commsense}. \hl{This paper, building on the CommSense framework, focuses on enhancing NLP capabilities using LLMs and testing the feasibility of integrating LLMs into such evaluation systems.}

\begin{table}[t!]
\centering
\caption{Communication metrics and their operational rules systematically identified for evaluating palliative care patient-provider communication. (This table is referred from \cite{lebaron2023commsense})}
\label{tab:criteria-commsense}
\footnotesize 
\begin{tabularx}{\textwidth}{l *{5}{X}}
\rowcolor[gray]{0.9} 
\hline
 & \textbf{Understanding} & \textbf{Empathy} & \textbf{Emotion} & \textbf{Presence} & \textbf{Clarity} \\  \hline
\textbf{Good} & Provider asks open-ended questions which invite the patient’s perspective. & Provider acknowledges the patient’s feelings; expresses support and opportunity for patient to express emotions. & Provider responds to patient’s emotion with emotional alignment/ acknowledgment; paused for 10 seconds after delivering difficult news. & Provider acknowledges based on active listening; paraphrases the patient's concerns; encourages silence and pauses. & Provider intentionally avoids using or explains difficult medical jargon; contextualizes next care step. \\ \hline
\textbf{Bad} & Not applicable & Not applicable & Provider's responds to patient's emotion with cognitive facts/figures; lacks emotional alignment/acknowledgment. & Provider frequently interrupts; dominates conversation without allowing for pauses. & Provider overuses medical jargon without explanation; constructs long, complicated sentences. \\
\bottomrule
\end{tabularx}
\end{table}

\subsection{Large Language Models in Health Care and Communication}

The advent of transformer-based models like BERT~\cite{devlin2019bert} and GPT~\cite{radford2018improving} has paved the way for more advanced iterations such as GPT-4~\cite{openai2023gpt4} and LLaMA-2~\cite{touvron2023llama}, marking significant progress in LLM capabilities, including advanced in-context learning and reasoning \cite{hahn2023theory}. \hl{Techniques like prompt engineering, particularly Chain-of-Thought (CoT) prompting introduced by Wei et al. \textit{et al.} \cite{wei2023chainofthought}, enabling LLMs to break down complex reasoning tasks into multiple steps and `think' step by step, have enhanced these models' reasoning abilities by guiding them through a sequence of intermediary steps. Wang \textit{et al.} \cite{wang2023selfconsistency} further refined this approach by incorporating diverse reasoning paths and majority voting to improve response consistency.}

In healthcare communication, LLMs have shown promise in areas like sentiment analysis and emotion detection, demonstrating their potential to understand nuanced human emotions  \cite{Koco__2023, qin2023chatgpt, zhong2023chatgpt}. For instance, studies \cite{lamichhane2023evaluation,amin2023affective,yang2023interpretable} have showcased the effectiveness of ChatGPT (GPT-3.5) in diagnosing mental health conditions such as stress and depression. Additionally, the development of Emotional Chain-of-Thought (ECoT) \cite{li2024enhancing} has significantly improved LLMs' ability to generate emotionally intelligent responses, closely mirroring human emotional understanding.

Building on these advancements, our research seeks to leverage LLMs to enhance patient-provider communication, a critical yet underexplored area of healthcare AI applications. With advanced audio transcription and speaker diarization tools like Whisper\footnote{Introducing Whisper. \url{https://openai.com/research/whisper}} and Otter.ai\footnote{Otter.ai. \url{https://otter.ai}}, now it is feasible to analyze linguistic context in clinical settings in real-time. This study pilot-tests the capability of LLMs to evaluate healthcare providers' communication skills using the content of clinical interactions.

\section{Feasibility Test of LLMs in Identifying Clinical Communication Metrics} \label{sec:benchmark}

Building on the foundation laid in the above sections, this section explores the feasibility of leveraging LLMs to identify critical metrics in clinical communication.

To address the challenges of sensitive clinical data and limited dialogue datasets, we established a set of benchmark scripts for evaluation (Section \ref{sec:generation}). These scripts were used to test LLMs including GPT-3.5, GPT-4, and Llama2-13b, using varying prompting strategies to enhance reasoning (Section \ref{sec:eval_approach}). We then assessed model performance both quantitatively (Section \ref{sec:eval_outcomes}) and qualitatively (Section \ref{sec:qual}).

\subsection{Generation of Clinical Communication Benchmark Scripts} \label{sec:generation}

Given the inherent sensitivity of clinical data and the notable lack of openly available clinical dialogue datasets, it became imperative to construct a series of benchmark scripts to facilitate this study. These scripts, referred to as "Benchmark Conversation Scripts\footnote{The scripts are available at \url{https://github.com/BarnesLab/PALLM/tree/main/scripts}}," were drafted by our clinical team to simulate real-world clinical interactions and were labeled with predefined communication metrics in Table \ref{tab:criteria-commsense} (i.e., understanding, empathy, emotion, presence, and clarity). The objective was to simulate realistic clinical dialogues and assess LLMs' capability to accurately identify these communicative metrics.

To this end, the clinical team conceived two distinct script categories tailored to physicians and nurses, reflecting the clinical roles of the human data generators. These categories focused on discussing pain management in advanced cancer scenarios and prognosis and goals of care. Each category (physician and nurse) comprised two `Good' scripts, exemplifying good communication practices such as the demonstration of empathy, and two `Bad' scripts, illustrating bad communication practices like the excessive use of medical terminology and conversational interruptions. A total of eight scripts were drafted by three clinical faculty members and subsequently annotated with the desired metrics by two separate clinical team members using Microsoft Word for independent labelling. Any discrepancies in annotation were resolved by group discussion to reach consensus. In order to capture `good' or `bad' communication elements, portions of conversations turns (referred to as `Segment') were tagged as either `good' (in `good' scripts) or `bad' (in `bad' scripts). For example, following the criteria listed in Table \ref{tab:criteria-commsense}, an annotation tagged as ``Emotion–Bad'' represented deflecting or dismissing patient emotion, whereas ``Emotion-Good'' represented responding therapeutically to patient emotion.

\subsection{Evaluation Approach} \label{sec:eval_approach}

Using the dataset crafted and annotated by healthcare professionals, we tested the ability of LLMs to identify communication metrics according to operational rules in Table \ref{tab:criteria-commsense}. Our approach involves providing the LLMs with the operational rules and prompting them to make decisions on a segment-by-segment basis for a given conversation. We employed three state-of-the-art LLMs: GPT-3.5, GPT-4, and the Llama2-13b version. We accessed the GPT models through the OpenAI API, while the Llama2-13b model is hosted on a local server equipped with an Nvidia A6000 GPU featuring 48GB of Graphics RAM. This section elucidates the design of our prompting methods and the underlying rationale behind their conception.

Note, throughout the current work, the evaluation employs binary classification tasks to analyze segments in either `Good' or `Bad' simulated scripts. That said, in `Good' scripts, segments are classified as `Good' or `None' (which means the rules did not apply), and similarly, in `Bad' scripts, as `Bad' or `None.' \hl{This binary classification serves as a preliminary step to assess feasibility, though we acknowledge real-world clinical interactions often involve a more complex spectrum of communication quality beyond simple binary categories.}

\begin{figure}
    \centering
    \includegraphics[width=0.83\linewidth]{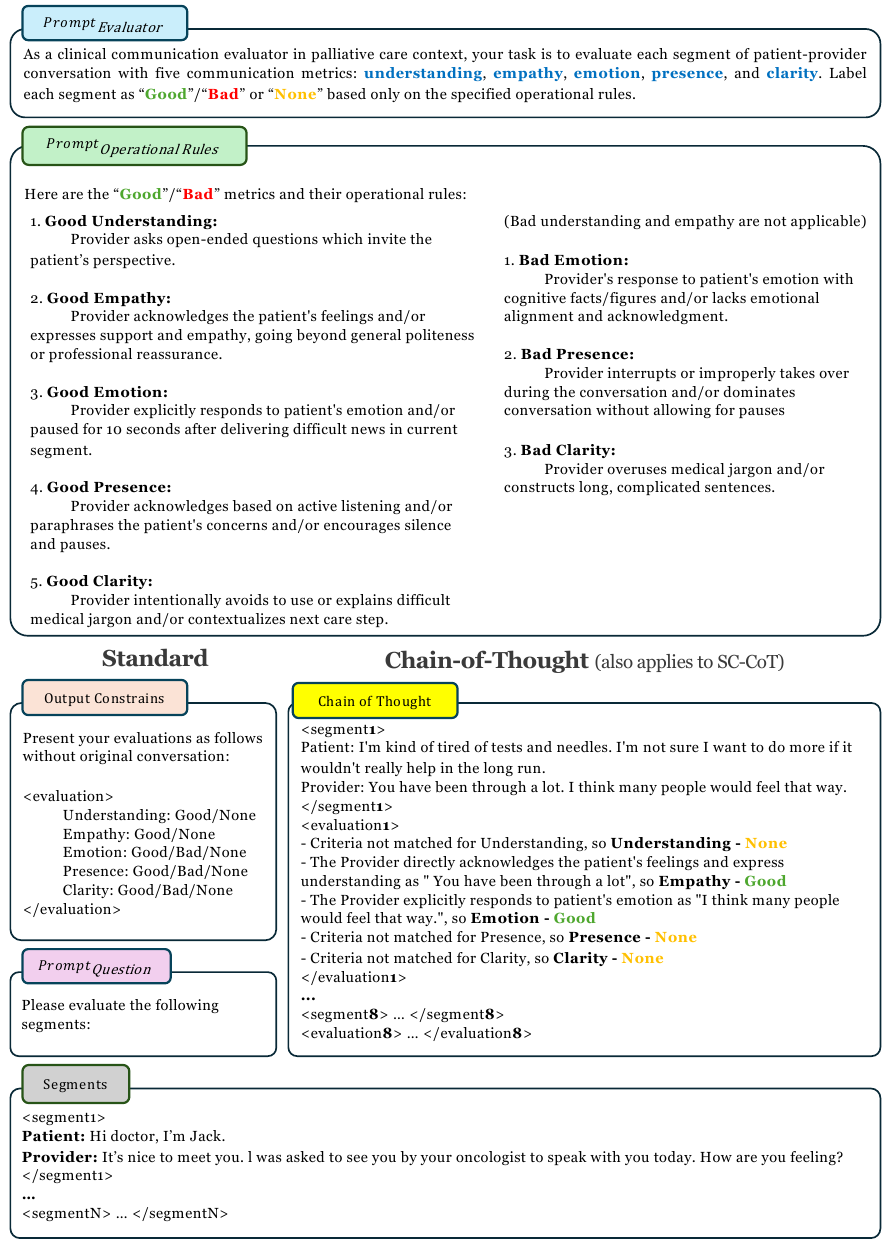}
    \caption{Prompt design of patient-provider communication evaluation in palliative care.}
    \label{fig:prompt_eval}
\end{figure}

Our prompt design aims to customize a pre-trained, general-purpose LLM with the capability to assess communication metrics per defined operational rules. Starting with a zero-shot prompt comprising solely the rules, we test the innate rule-application ability of the models. \hl{To enhance the LLMs' reasoning capabilities, we then incorporate in-context exemplars that demonstrate correct judgments and reasoning processes into more advanced ``chain-of-thought'' (CoT) \cite{wei2022chain} and ensemble ``self-consistency CoT'' \cite{wang2023selfconsistency} prompts.} Here are the details of the prompting approaches:

\subsubsection{Standard Prompt Structure:} We propose the following content factors to be integrated as a complete prompt (see Figure \ref{fig:prompt_eval}'s left side):
\begin{equation*}
    Prompt_{\textbf{standard}} = {Prompt}_\textbf{evaluator} + {Prompt}_\textbf{operational\_rules} + \text{Output\_Constraint} + {Prompt}_\textbf{question} + \text{Segments}
\end{equation*} where
\begin{itemize}
    \item[1)] ${Prompt}_\textbf{evaluator}$ initiates the sequence, establishing the LLM's role as a clinical communication evaluator;
    \item[2)] ${Prompt}_\textbf{operational\_rules}$ provides a detailed enumeration of the operational rules for evaluation, formatted in comprehensible language;
    \item[3)] $OutputConstraint$ specifies the expected format of the LLM's response, guiding it towards definitive outcomes, i.e., `Good', `Bad', or `None';
    \item[4)] ${Prompt}_\textbf{question}$ poses the specific evaluation query to the LLM;
    \item[5)] Finally, the patient-provider conversation segment(s)\footnote{The number of segments prompted can vary among different LLMs, influenced by each model's maximum token handling capacity.} for assessment are attached.   
\end{itemize}

\subsubsection{Enhanced Reasoning Prompts:} To further enhance the LLMs' reasoning ability and alignment with the operational rules, we apply two advanced prompting strategies:

\begin{itemize}
    \item \textbf{Chain-of-Thought (CoT) \cite{wei2022chain}:} This approach guides the LLM through a human-like, step-by-step reasoning process. Instead of using ${Prompt}_\textbf{question}$ and $OutputConstraint$ to lead to question answering, $Prompt_{CoT}$ includes several conversation segment exemplars (in our case, we manually generated eight new exemplars respectively for both `Good' instances and `Bad' instances covering all operational rules. The generated exemplars are independent of the eight benchmark scripts). Each exemplar includes a dialogue segment, the reasoning process for each metric according to the operational rules, and the final decision derived from this reasoning. The detailed prompt design (also see Figure \ref{fig:prompt_eval}'s right side):
    \begin{equation*}
    Prompt_{\textbf{CoT}} = {Prompt}_\textbf{evaluator} + {Prompt}_\textbf{operational\_rules} + \text{CoT Exemplars} + \text{Segments}
\end{equation*}    
    \item \textbf{Self-Consistency Chain-of-Thought (SC-CoT) \cite{wang2022self}:} This method generates multiple CoT reasoning paths and selects the most consistent response as final decision, enhancing the reliability of the evaluation.
\end{itemize}

\hl{These three strategies are designed to improve the LLMs’ reasoning process. In other words, beyond standard prompts, CoT guides the LLMs to reason step by step, similar to a human evaluator's thought process. SC-CoT aims to further strengthen this by using diverse CoT prompts for the same task, ultimately improving the final judgment through an ensemble approach, similar to the integration of multiple evaluators to enhance robustness.}

The subsequent evaluation of the models utilizes Balanced Accuracy (adjusted accuracy to account for class imbalance), Precision, and Recall, offer a comprehensive view of their performance. \hl{Specifically, healthcare professionals annotated each provider's turn of speech as `Good', `Bad', or `None' (where no good or bad metrics applied), and these labels were compared with the LLMs' annotations, which followed the same labeling criteria. We conducted a turn-by-turn comparison of the LLMs' annotations with human expert judgments across all eight scripts.}

Additionally, we compare the LLM-based methods with Wang et al. \cite{wang2024commsense}, which could serve as a benchmark for non-LLM NLP methods in evaluating clinical communication. Specifically, \cite{wang2024commsense} presents a computational framework that uses a mix of existing NLP techniques in a rule-based approach for the same evaluation tasks. For instance, as `Good Understanding' was defined as `the provider asked open-ended questions which invites the patient’s perspective.' To identify this, Wang et al. \cite{wang2024commsense} trained a BERT model \cite{devlin2018bert} based on Amazon Review Dataset \cite{he2016ups} to detect open-ended questions. When this model finds a provider’s question that is open-ended, it would be annotated as `Understanding-Good'. 

\subsection{Evaluation Results} \label{sec:eval_outcomes}

The evaluation results of rule-based NLP non-LLM baseline and a range of LLMs, i.e., LLaMa2-13b, GPT-3.5, and GPT-4, under different prompting strategies are presented in Table \ref{tab:evaluation}. Examples of GPT-4’s outputs under different prompting strategies are shown in Appendix I, Figure \ref{fig:evaluation_gpt4}. The results showcase that GPT-4, which has the largest model parameter count, consistently outperforms the other models across all metrics. Compared to the non-LLM benchmark, the LLMs generally exceed the benchmark in most metrics, indicating their advanced performance in evaluating clinical communication. However, LLaMa2-13b, with its smallest parameter count of 13 billion (compared to 175 billion and 1.8 trillion), struggled to effectively identify most communication metrics, often performing near the random guess baseline of 50\% in most cases.

When comparing the three prompting methods, the Standard strategy showed variable model performance, with GPT-4 consistently outperforming both LLaMa2-13b and GPT-3.5 across all metrics. This pattern was also observed in evaluations of `Bad' scripts. The CoT method significantly improved GPT-4’s performance, particularly with `Good' scripts, where it achieved the highest metrics, indicating strong alignment with operational rules and enhanced reasoning capabilities. The SC-CoT strategy further boosted GPT-4’s performance, especially in `Good' script evaluations, demonstrating that consistency in reasoning paths significantly increases the reliability of the model's communication metric assessments.

Across the five communication metrics, GPT-4, when using CoT and SC-CoT strategies, generally performed comparably to human labeling, with over 90\% accuracy in understanding and empathy for `Good' scripts, and over 95\% accuracy in clarity, as well as in `Bad' presence and clarity. However, there is room for improvement in accurately capturing emotion and presence metrics, particularly with `Good' scripts.

\begin{table}[t!]
    \centering
    \caption{Results of balanced accuracy, precision, and recall across rule-based NLP methods and various-scale LLMs: LlaMa2 (13-billion), GPT-3.5 (175-billion), and GPT-4 (approximately 1.8 trillion) using different prompting strategies (standard, CoT, and SC-CoT). We emphasize \textbf{balanced accuracy} as the primary metric for assessing overall model performance while also presenting precision and recall to offer a comprehensive view of the model behaviors. }
    \label{tab:evaluation}
    {\resizebox{\textwidth}{!}{
    \begin{tabular}{lccccccc}
    \toprule
    \multirow{2}{*}{\textbf{Method}} & \multirow{2}{*}{\textbf{Script}} & \multirow{2}{*}{\textbf{Model}} & \multicolumn{5}{c}{\textbf{Balanced Accuracy} / Precision / Recall (\%)} \\
    \cmidrule{4-8}
     & &  & Understanding & Empathy & Emotion & Presence & Clarity \\
     \midrule
    \textbf{Non-LLM} & \text{Good} & \text{Mix} & 74.47/72.95/78.98 & 65.00/67.79/66.20 & \textbf{85.09}/73.45/84.57 & 82.15/72.28/70.21 & 76.56/61.39/56.72 \\
    \cmidrule{2-8}
    \textbf{Benchmark} \cite{wang2024commsense}& \text{Bad} & \text{Mix} & - & - & 71.87/71.98/67.82 & 78.69/83.47/63.78 & 78.45/65.22/81.25\\
    \midrule
     & \multirow{3}{*}{Good} & \text{LLaMA2-13b} & 55.00/55.00/100.00 & 51.26/40.35/92.00 & 55.21/25.86/100.00 & 55.88/21.05/100.00 & 53.85/18.64/100.00 \\
    & & \text{GPT-3.5} & 54.24/71.43/15.15 & 74.52/86.67/54.17 & 58.67/60.00/21.43 & 67.92/75.00/40.00 & 74.02/85.71/50.00\\
    \textbf{Standard} & & \text{GPT-4} & 82.42/82.35/84.85 & 86.54/83.33/83.33 & 70.41/85.71/42.86 & 73.75/60.00/60.00 & 81.37/80.00/66.67\\
    \cmidrule{2-8}
    \textbf{Prompting}& \multirow{3}{*}{Bad} & \text{LLaMA2-13b} & - & - & 67.01/66.67/77.78 & 71.33/50.00/66.67 & 52.96/36.36/36.36\\
    & & \text{GPT-3.5} & - & - & 62.85/72.73/44.44 & 87.50/100.00/75.00 & 93.28/90.91/90.91\\
    & & \text{GPT-4} & - & - & 77.08/85.71/66.67 & 87.98/70.00/87.50 & 95.45/100.00/90.91\\
    \midrule
     & \multirow{3}{*}{Good} & \text{LLaMA2-13b} & 61.61/61.70/85.29 & 52.47/41.30/76.00 & 52.29/25.00/73.33 & 62.62/38.78/95.00 & 58.22/20.93/81.82 \\
    & & \text{GPT-3.5} & 71.97/80.00/60.61 & 73.24/81.25/54.17 & 57.65/50.00/21.43 & 66.46/40.91/60.00 & 82.52/66.70/72.73\\
    \textbf{Chain-of-} & & \text{GPT-4} & 83.94/82.86/87.88 & 90.38/95.24/83.33 & 72.45/57.14/57.14 & 78.12/66.70/66.67 & 94.23/64.71/100.00\\
    \cmidrule{2-8}
    \textbf{Thought} \cite{wei2022chain} & \multirow{3}{*}{Bad} & \text{LLaMA2-13b} & - & - & 76.04/75.00/80.33 & 70.44/40.00/88.89 & 73.52/52.94/81.82 \\
    & & \text{GPT-3.5} & - & - & 62.85/72.73/44.44 & 84.62/50.00/100.00 & 75.89/52.63/90.91\\
    & & \text{GPT-4} & - & - & 82.89/60.00/83.33 & 90.76/63.64/87.50 & \textbf{99.21}/92.31/100.00 \\
    \midrule
     & \multirow{3}{*}{Good} & \text{LLaMA2-13b} & 60.14/60.87/82.35 & 52.47/41.30/76.00 & 48.96/23.26/66.67 & 62.62/38.78/95.00 & 58.22/20.93/81.82 \\
    \textbf{Self-} & & \text{GPT-3.5} & 60.15/76.92/30.30 & 56.63/42.86/21.43 & 56.63/42.86/21.43 & 64.79/54.55/40.00 & 88.02/75.00/81.82\\
    \textbf{Consistency}  & & \text{GPT-4} & \textbf{90.45}/90.91/90.91 & \textbf{93.75}/100.00/87.50 & 75.51/72.73/57.14 & \textbf{84.58}/84.62/73.33 & \textbf{96.15}/73.33/100.00\\
    \cmidrule{2-8}
    \textbf{Chain-of-}& \multirow{3}{*}{Bad} & \text{LLaMA2-13b} & - & - & 76.04/75.00/83.33 & 70.44/40.00/88.89 & 73.52/52.94/81.82 \\
    \textbf{Thought} \cite{wang2022self}& & \text{GPT-3.5} & - & - & 63.19/77.78/38.89 & 90.38/61.54/100.00 & 78.06/55.56/90.91\\
    & & \text{GPT-4} & - & - & \textbf{87.85}/85.00/94.44 & \textbf{96.15}/80.00/100.00 & 95.45/100.00/90.91\\
    \bottomrule
    \end{tabular}}}
\end{table}

\subsection{Qualitative Notes} \label{sec:qual}

We manually checked and analyzed the responses from different models under various prompting setups, focusing on the subtleties of their outputs. This section summarizes the observed model behaviors and outcomes.

Using standard prompting, we asked the LLMs to explain their reasoning after evaluating the segments to better understand their justifications. We found that the models often created incorrect operational rules, especially when identifying traits as ``Good'' in the communication metrics, leading to many false positives where segments were wrongly classified as ``Good'' despite not meeting the criteria.

\hl{Additionally, even when instructed to remain impartial, LLMs tend to provide more positive assessments. This bias likely stems from their training, which emphasizes producing outputs that are less harmful and more agreeable, aligning with general human preferences \cite{Ji2023Controlling}. Additionally, we acknowledge and highlight potential biases related to demographics, diseases, and clinical settings in the model's outputs warrant further investigation.} This tendency was especially strong in the smaller LlaMa2-13b model, which showed a greater inclination to assign more `Good' labels and fewer `Bad' labels.

Our review revealed that the LlaMa2-13b model \hl{often struggled with reasoning, frequently offering} irrational justifications. However, applying CoT prompting significantly improved its reasoning and decision making. 

\hl{Manual check confirmed that CoT-enhanced LLMs, such as GPT-4, produced mostly accurate and logically sound reasoning.} The CoT approach also improved GPT-4's adherence to operational guidelines, reducing overly positive judgments. Overall, CoT method enhanced the alignment of all LLMs with the established operational rules.

\section{Fine-Tuning LLMs To Be Task-Specific for Clinical Communication Evaluation}

In this section, we fine-tune (a method of adapting a pre-trained model to a specific dataset to improve its performance on a particular task \cite{peters2019tune}) the open-source LLaMa-2-13b model. \hl{Our goal is to improve the model’s ability to evaluate clinical communication while leveraging LLaMa-2-13b’s flexibility and privacy-preserving features, making it suitable for institutional deployment.}

\hl{We take advantage of GPT-4’s superior capabilities, as demonstrated in Section \ref{sec:benchmark}, to generate synthetic datasets that help fine-tune LLaMa-2-13b. This approach allows the model to become task-specific with fewer parameters than GPT-4, demonstrating the potential of small-scale LLMs for specialized tasks while enabling flexible in-house LLM deployment and minimizing computational resource demands.}

\subsection{Generating Synthetic Conversation Evaluation Datasets using GPT-4}

We use GPT-4 to create synthetic datasets for evaluating patient-provider conversations, aiming to provide a comprehensive, diverse, and realistic collection for training the LLaMa2-13b model in clinical communication assessment. Building on insights from \cite{brown2020language} about LLMs' capability to produce quality output from few-shot examples, and supported by \cite{puri2020training,hamalainen2023evaluating,yu2024large}, which highlight the potential of LLMs in generating synthetic data to overcome and limitations of human-subject data collection, we use GPT-4 to create detailed patient-provider dialogue segments and their associated labels (and reasoning process) for communication metrics.

The dataset generation process follows the practices suggested by \cite{chintagunta2021medically,peng2023study,hamalainen2023evaluating,tang2023does} for generating synthetic datasets in medical, clinical text, and human-computer interaction domains. This process involves two key steps: 1) diversified data generation and 2) quality assurance, resulting in 3,000 diversified segments of palliative clinical communication.

\begin{figure}
    \centering
    \includegraphics[width=0.85\linewidth]{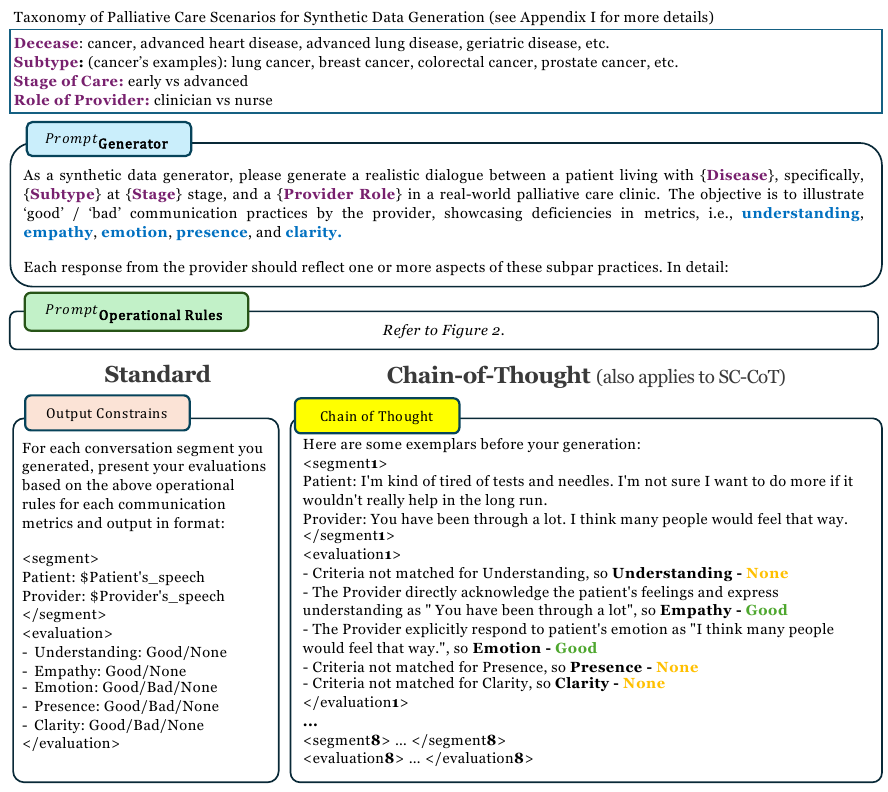}
    \caption{Prompt design for diversified and realistic synthetic dataset of palliative care clinical communication.}
    \label{fig:generate_segment}
\end{figure}

\textbf{Diversified Data Generation:} We used GPT-4 to generate patient-provider conversation samples, each starting with a patient’s statement followed by a healthcare provider’s response. To ensure diversity and realism, we developed a taxonomy of palliative care scenarios to guide the data generation. Based on \cite{ewert2016building,wikert2022specialist},  this taxonomy categorizes scenarios according to various provider roles (e.g., clinicians and nurses), patient care stages (early to advanced), and a range of severe diseases, including their subtypes (e.g., various cancers, advanced heart disease, and neurological and geriatric conditions). The aim is to create a dataset that accurately reflects the diverse spectrum of patient-provider interactions in palliative care, capturing the nuances and complexities of real-world scenarios. Figure \ref{fig:generate_segment} illustrates the prompting workflow used to generate these patient-provider conversation samples based on the taxonomy.

\textbf{Standard Prompt Structure:} We design the following components into $Prompt_{\textbf{generation}}$ to generate synthetic dataset (see Figure \ref{fig:generate_segment}'s left side):
\begin{equation*}
    Prompt_{\textbf{generation}} = {Prompt}_\textbf{generator} + {Prompt}_\textbf{operational\_rules} + \text{Output\_Constrains}
\end{equation*} where
\begin{itemize}
    \item[1)] ${Prompt}_\textbf{generator}$ initiates the sequence, establishing the LLM's task to generate synthetic dialogue;
    \item[2)] ${Prompt}_\textbf{operational\_rules}$ provides a detailed enumeration of the operational rules for evaluation, formatted in comprehensible language;
    \item[3)] $OutputConstraint$ specifies the expected format of the LLM's response, guiding it towards definitive outcomes, i.e., `Good', `Bad', or `None';
\end{itemize}

See Figure \ref{fig:generation_standard} in Appendix II for example outputs generated under $Prompt_{\textbf{generation}}$.

\textbf{CoT Prompt Structure:} $Prompt_{generation\_CoT}$ includes several conversation segment exemplars (in our case, we manually generated eight new exemplars respectively for both `Good' instances and `Bad' instances covering all operational rules. (see Figure \ref{fig:generate_segment}'s right side):
\begin{equation*}
    Prompt_{\textbf{generation\_CoT}} = {Prompt}_\textbf{generator} + {Prompt}_\textbf{operational\_rules} + \text{CoT Exemplars}
\end{equation*} where
\begin{itemize}
    \item[1)] ${Prompt}_\textbf{generator}$ initiates the sequence, establishing the LLM's task to generate synthetic dialogue;
    \item[2)] ${Prompt}_\textbf{operational\_rules}$ provides a detailed enumeration of the operational rules for evaluation, formatted in comprehensible language;
    \item[3)] $\text{CoT Exemplars}$ guides the LLM through a step-by-step reasoning process, akin to human problem-solving methodologies towards definitive outcomes, i.e., `Good', `Bad', or `None';
\end{itemize}

See Figure \ref{fig:generation_cot} in Appendix II for example outputs generated under $Prompt_{\textbf{generation\_CoT}}$.


\subsection{Parameter-Efficient Fine-Tuning of Llama2-13b}

This subsection delves into the fine-tuning of Llama2-13b, using Parameter-Efficient Fine-Tuning (PEFT) \cite{peft} techniques, notably Low-Rank Adaptation (LoRA) \cite{hu2021lora}, to enhance the model’s performance in clinical communication evaluation. \hl{We fine-tuned a selective subset of Llama2-13b’s parameters with the synthetic dataset, in order to tailor and optimize Llama2-13b’s task-specific ability to analyze patient-provider dialogues with its smaller parameter scale compared to GPT-3.5 and GPT-4 models.}

Specifically, by utilizing LoRA, we adjust the model’s low-rank weight components, focusing on the most influential features for output generation. Our fine-tuning employs a rank of 8 (capturing only the top 8 features in the weight matrices) and an alpha parameter of 32 (scaling the adjustments to low-rank components), resulting in the training of 6,553,600 parameters—just 0.0503\% of Llama2-13b’s total—over 50 epochs.

\subsection{Fine-Tuning Outcomes} \label{sec:fine-tune-outcome}

\hl{The fine-tuning of the Llama2-13b model yielded significant improvements compared to its original capacity presented in Section \ref{sec:benchmark} (50\% balanced accuracy in most cases).} According to Table \ref{tab:fine-tune}, balanced accuracy scores are segmented based on the number of training samples and the prompting method employed, providing a comprehensive view of the fine-tuning's effectiveness. We also attached fine-tuned Llama2-13b's precision and recall scores under the same settings in Appendix III's Table \ref{tab:fine-tune-all}.

\textbf{Impact of Training Sample Size:} The results indicate a clear trend that increasing the number of synthetic training samples improves the model's performance across all metrics. For instance, when using the SC-CoT method, the balanced accuracy in `Emotion-Good' improves from 48.96\% with 1000 samples to 84.69\% with 3000 samples. This trend is consistent across other metrics and methods, emphasizing the importance of the dataset's size in fine-tuning processes.

\textbf{Prompting Insights:} The table also compares different prompting methods, i.e., standard, CoT, and SC-CoT. Notably, the CoT and SC-CoT methods generally outperform the standard method, especially in the larger sample sizes. For example, in the `good' script category with 3000 samples, the SC-CoT method achieves a remarkable 85.21\% in `presence' and 87.76\% in `clarity', showcasing the effectiveness of these advanced prompting strategies in enhancing the model’s understanding and evaluation capabilities.

\begin{table}[t!]
    \centering
    \caption{Results of \textbf{fine-tuned Llama2-13b} models' balanced accuracy across all communication metrics within the script type. The Llama2-13b model was fine-tuned based on varying number of synthetic data samples generated under different prompting methods.}
    \label{tab:fine-tune}
    {\resizebox{0.65\textwidth}{!}{
    \begin{tabular}{lccccccc}
    \toprule
    \multirow{2}{*}{\textbf{Method}} & \multirow{2}{*}{\textbf{Training sample}} & \multirow{2}{*}{\textbf{Script}} & \multicolumn{5}{c}{\textbf{Balanced Accuracy (\%)}} \\
    \cmidrule{4-8}
     & &  & Understanding & Empathy & Emotion & Presence & Clarity \\
     \midrule
    \multirow{6}{*}{\textbf{Standard}} & \multirow{2}{*}{1000} & \text{Good} & 54.24 & 55.21 & 58.67 & 54.17 & 63.48\\
    & & \text{Bad} & - & - & 67.01 & 71.33 & 52.96\\
    \cmidrule{2-8}
    & \multirow{2}{*}{2000} & \text{Good} & 55.00 & 51.26 & 65.31 & 55.88 & 53.85 \\
    & & \text{Bad} & - & - & 79.86 & 78.37 & 60.28\\
    \cmidrule{2-8}
    & \multirow{2}{*}{3000} & \text{Good} & 56.82 & 66.35 & 74.52 & 67.92 & 74.02\\
    & & \text{Bad} & - & - & 79.86 & 78.37& 60.28\\
    \midrule
    \multirow{6}{*}{\textbf{CoT}} & \multirow{2}{*}{1000} & \text{Good} & 61.61 & 52.47 & 52.29 & 62.62 & 58.22 \\
    & & \text{Bad} & - & - & 65.28& 71.33 & 70.95\\
    \cmidrule{2-8}
    & \multirow{2}{*}{2000} & \text{Good} & 65.61 & 69.71 & 73.12 & 56.94 & 83.92 \\
    & & \text{Bad} & - & - & 73.96 & 67.33 & 86.76\\
    \cmidrule{2-8}
    & \multirow{2}{*}{3000} & \text{Good} & 68.48 & 71.79 & 81.12 & 79.79 & \textbf{87.76}\\
    & & \text{Bad} & - & - & \textbf{80.56} & \textbf{82.21} & \textbf{82.02}\\
    \midrule
    \multirow{6}{*}{\textbf{SC-CoT}} & \multirow{2}{*}{1000} & \text{Good} & 60.14 & 52.47 & 48.96 & 62.62 & 58.22 \\
    & & \text{Bad} & - & - & 65.28 & 76.89 & 70.95 \\
    \cmidrule{2-8}
    & \multirow{2}{*}{2000} & \text{Good} & 68.64 & 70.51& 77.29 & 59.21& 84.88 \\
    & & \text{Bad} & - & - & 71.88 & 52.67 & 79.64 \\
    \cmidrule{2-8}
    & \multirow{2}{*}{3000} & \text{Good} & \textbf{70.15} & \textbf{74.84}& \textbf{84.69} & \textbf{85.21} & \textbf{87.76} \\
    & & \text{Bad} & - & - & \textbf{80.56} & \textbf{82.21} & \textbf{82.02} \\
        \bottomrule
    \end{tabular}}}
\end{table}

\section{Discussion}

In this section, we discuss the findings, implications, and ethical and privacy considerations of this work.

\subsection{Summary of Findings}

Here we summarize the findings of this work: 

\textbf{LLMs exhibit state-of-the-art performance in evaluating clinical conversation.} The evaluation outcomes underscore the state-of-the-art performance of LLMs in the realm of clinical communication assessment. Notably, GPT-4 has demonstrated exceptional performance, consistently surpassing existing non-LLM benchmarks, GPT-3.5 and Llama2-13b, across nearly all communication metrics. Furthermore, this study illustrates the incremental but notable improvement in LLMs performance when subjected to prompting methods for reasoning enhancement. This leap in performance demonstrates the feasibility of LLMs for nuanced and accurate evaluation of clinical communication, setting a new benchmark in the field.

\textbf{LLMs Facilitate Personalized Feedback Through Chain-of-Thought Reasoning.} The study utilized LLMs, either through prompting with CoT examples (as discussed in Section \ref{sec:eval_outcomes}) or by fine-tuning (outlined in Section \ref{sec:fine-tune-outcome}), to enhance their reasoning capabilities. \hl{This approach not only improves the models’ performance but also generates chain-of-thought outputs providing straightforward and logical reason why the judgement was made, explaining strengths or areas for improvement. For instance, LLMs with CoT can pinpoint the use of specific medical jargon and could offer suggested explanations to the provider (e.g., the model outputted: ``\textit{The provider used the term `hypertension', which is jargon, so clarity is compromised. This term can be replaced with `high blood pressure'.}"), thereby facilitating personalized and constructive feedback.}

\textbf{Smaller-Scale LLMs Show Promise with Synthetic Data Fine-Tuning for Task-Specific \hl{In-House Deployment.}} \hl{The fine-tuned Llama2-13b model demonstrated significant accuracy improvements over the baseline across all evaluated metrics, surpassing the performance of the GPT-3.5 model (with 175 billion parameters) despite being 13.46 times smaller.} It also achieved comparable results to GPT-4, which is 138.46 times larger (1.8 trillion parameters). This highlights the potential of smaller, task-specific LLMs fine-tuned with synthetic or real-world data for clinical communication analysis. \hl{Such models offer a practical solution for deployment in environments where in-house hosting is essential, providing a flexible approach to integrating LLMs into clinical settings.}

\subsection{Clinical Implications}

Effective patient-provider communication is crucial for patient satisfaction and optimal health outcomes. Our research highlights the potential of LLMs to enhance communication in palliative care, where empathy and understanding are essential. This work marks a significant step toward integrating LLMs into clinical care communication, particularly in delivering responses that demonstrate empathy and understanding, which are critical in palliative care. The integration of LLMs aims to improve healthcare providers' communication skills and enhance patient outcomes by offering feedback in various formats: real-time alerts (e.g., vibrations for excessive medical jargon), post-conversation reviews (highlighting strengths and areas for improvement), and longitudinal evaluations that track progress over time.

Incorporating LLMs into communication training and quality improvement for clinicians represents progress in elevating healthcare interactions. By providing personalized feedback, \hl{LLMs can help refine essential communication skills}, such as building rapport with patients, without disrupting clinical workflows or affecting providers' diagnostic or treatment decisions. These advancements hold promise not only for improving individual provider interactions but also for enhancing communication standards within healthcare systems.

The role of LLMs in other clinical practices and educational scenarios, such as telehealth \cite{henry2017clinician}, online text communication \cite{wang2024rapport}, and healthcare team communication \cite{pulman2009enabling}, offers significant opportunities for further research. Adapting LLM-driven feedback for virtual consultations or asynchronous digital exchanges can strengthen communication rapport, fostering a strong therapeutic alliance and improving patient outcomes. This application of LLMs has the potential to enhance patient care and create safer healthcare environments.

In summary, our pilot study lays the foundation for future research and application of LLMs in clinical communication settings. Exploring LLMs across varying healthcare contexts, from clinical encounters to telehealth, presents a promising opportunity to improve the quality of healthcare communication.

\subsection{\hl{Ethical and Privacy Considerations}}

This study builds upon a broader research project exploring the feasibility and acceptability of using ubiquitous wearables and NLP approaches for scalable, non-invasive, and ethical clinical communication evaluation \cite{lebaron2022exploring, lebaron2023commsense, wang2024commsense}. The project included audio recording and transcription via smartwatches, linguistic analysis, evaluation, and the provision of constructive feedback. In this initial proof-of-concept study, 40 healthcare professionals and trainees participated in eight scripted scenarios in a controlled lab environment, with approval from the University of Virginia Social and Behavioral Sciences Institutional Review Board (UVA SBS IRB \#4985). This PALLM work focused on a preliminary examination of LLMs' linguistic analysis capabilities, only using the scripted scenarios as test cases. \hl{As we advance the research and deployment of an LLM-powered communication evaluation system, prioritizing ethical, privacy, and safety considerations is crucial given the sensitive nature of clinical interactions.}

\hl{The goal of PALLM -- to enhance healthcare providers’ communication skills, thereby improving patient health outcomes -- must carefully consider the implications and ethics of utilizing such technologies. Participants must be consented and adequately informed of the potential risks of such technology, and be fully aware of how their data will be used. }

\hl{Safeguarding the confidentiality of sensitive clinical communication data is crucial for the system's operation, ensuring compliance with standards such as The Health Insurance Portability and Accountability Act of 1996 (HIPAA) \cite{act1996health}. Importantly, this paper demonstrates the feasibility of developing in-house LLMs, eliminating the need to upload data to third-party providers, a critical strategy to enhance patient privacy and data security. We emphasize that future systems should maintain sensitive data within a secure, institution-controlled or HIPAA-compliant infrastructure, ensuring clinical data is not used to train models, thereby minimizing risks and enhancing data privacy and security.}

Additionally, the generalizability and adaptability of the system should be considered when deployed with different stakeholders and in different contexts. For example, the system should be adaptable to diverse clinical situations, such as collaborative team discussions or conversations with multiple speakers, such as those involving patients’ family members. It should also be inclusive of potentially marginalized patient groups, such as older individuals or those who are multilingual speakers. Moreover, communication evaluation results must be contextualized.  For example, the appropriateness of medical jargon usage may vary depending on patients’ backgrounds. While employing medical terminology might be suitable for patients with prior medical knowledge, it could be suboptimal for others lacking such a background.

\hl{Finally, while LLMs have shown exceptional potential in evaluating clinical communication, challenges remain for reliable deployment. For instance, false or biased judgments may place additional stress on healthcare providers. We call for the need for responsible benchmarking of LLMs’ performance, potential biases, and limitations across diverse clinical contexts, particularly with minoritized and underrepresented groups and in disadvantaged areas. Integrating LLMs with stakeholder-centered system designs, along with iterative refinements based on stakeholder feedback, is also essential for ethical clinical implementation.}

\section{Limitations and Future Work}

This section outlines the limitations of our study and charts a path for future research and engineering efforts to develop an actionable sensing and assistance system for real-world clinical deployment.

\subsection{Towards Real-World Data Collection and Clinical Validation}

Due to the lack of relevant accessible datasets on clinical dialogues, we created a small simulated benchmark dataset with input from clinicians in cancer care and pain management. This allowed us to contextualize and pilot-test LLMs in patient-provider communication encounters and generate a synthetic dataset for fine-tuning. \hl{While our foundational study has demonstrated the feasibility of using LLMs to evaluate clinical communication, it is clear that simulated datasets only scratch the surface of complex clinical interactions, reducing their effectiveness of evaluating real-world use case. Furthermore, content generated by LLMs may carry inherent biases [18], potentially impacting the validity and generalizability of the findings. The next critical step involves transitioning from these preliminary findings to real-world clinical data collection, where LLMs are rigorously tested in authentic, diverse, and multilingual clinical settings. This phase is essential to ensure that the models not only perform well in controlled scenarios but also adapt effectively to the complexities and variability of actual clinical interactions.}

\hl{Clinical validation is essential for the safe, effective, and ethical integration of AI-enhanced systems into clinical practice. However, recent research has revealed that between 2016 and 2022, only 56\% of medical AI devices and systems authorized by the FDA included reported clinical validation. The remaining authorizations were either validated retrospectively or prospectively, with 43.4\% lacking any reported clinical validation data \cite{elnot}. Particularly concerning is the lack of rigorous testing for fairness across diverse scenarios, especially in marginalized areas and among underrepresented groups. We emphasize the critical need for comprehensive benchmarking of model performance across different settings, thorough assessment of potential risks and biases, and the implementation of robust mitigation strategies. These steps are vital for ensuring responsible and trustworthy real-world clinical integration.}

\subsection{Towards Stakeholder-Centered, Ethical Clinical Integration}

Moreover, while studies \cite{sandhu2020integrating,lebaron2023commsense} have explored the feedback of licensed clinicians and pre-licensure medical and nursing students on technology-assisted education and evaluation, specific feedback on using GenAI and LLMs to analyze clinical communication performance remains underexplored. This study focused solely on assessing technological feasibility from a quantitative perspective. \hl{Future research should incorporate qualitative feedback before and after clinical integration from all stakeholders -- including patients, providers, engineers, administrators, educators, policymakers, and others -- to understand their insights and concerns towards a stakeholder-centered LLM-powered systems for healthcare communication.}

\hl{Future clinical integration must address the technical, ethical, and logistical challenges of integration. Key considerations include safeguarding data privacy, ensuring system interoperability, and providing comprehensive training for healthcare providers regarding the use of the technology. Developing and implementing this framework will require close collaboration across multiple disciplines (e.g., LLM engineering, nursing, medicine, human computer interaction, etc.) to fully harness the potential of LLMs to enhance clinical communication and improve patient outcomes. This stakeholder-centered consideration should be across the contextualization, development, validation, clinical integration, and long-term refinement of such AI-enhanced healthcare systems.  As the field of LLMs continues to evolve rapidly, we also call for the establishment of new benchmarks and best practices to ensure these technologies remain effective, ethical, and aligned with the needs of all stakeholders.}

\subsection{Multimodal and Adaptive Assessment and Feedback}

Clinical care communication is inherently multimodal, \hl{involving language, facial expressions, body language, and eye contact}, yet this work primarily focuses on the linguistic aspect. It is important to acknowledge that other elements of human interaction—such as body language, affective touch, physiological responses, and the clinical environment—are also crucial for a comprehensive understanding. \hl{Wearable devices like smart watches and glasses equipped with multimodal GenAI tools offer promising ways to capture and interpret these interactions. Initial studies have explored the use of wearable, wireless, and visual sensing technologies to better understand human interactions.} Research in ubiquitous sensing \cite{lebaron2023commsense,wang2023detecting,harari2023understanding,woods2024integrating}, ambient intelligence \cite{haque2020illuminating,krishna2022socially}, and multimodal GenAI \cite{mesko2023impact,ye2023mplug} has begun to investigate the perception and interpretation of human behavior and mental states across different contexts. However, there is still a significant gap in integrating these diverse inputs into a cohesive, patient-centered healthcare communication ecosystem. The future goal can be to combine these multimodal data streams to provide a holistic view of patient-provider interactions, enabling more nuanced and effective communication strategies that align with patient needs and preferences.

\section{Conclusion}

\hl{In summary, this study highlights the potential of LLMs in evaluating and enhancing clinical communication in palliative care. By using varying prompting techniques, LLMs outperformed traditional non-LLM NLP baselines in evaluating key metrics like empathy and understanding. Additionally, the fine-tuning of smaller models on synthetic datasets shows promise for practical deployment in healthcare settings. This research sets the stage for integrating LLMs into clinical communication workflows, aiming to improve patient-provider interactions and health outcomes. We also envision future efforts need to focus on real-world validation and ethical, stakeholder-centered integration of such technologies into the clinical environment.}

\section*{Acknowledgement}

This work was supported in part by the University of Virginia Engineering in Medicine Seed Pilot Program, the Gordon and Betty Moore Foundation, the National Institute of Mental Health of the National Institutes of Health under award number R01MH132138, and the Commonwealth Cyber Initiative, an investment in the advancement of cyber R\&D, innovation, and workforce development. For more information about CCI, visit www.cyberinitiative.org.

\bibliographystyle{ACM-Reference-Format}
\bibliography{reference}
\clearpage
\section*{Appendix}

\subsection*{I. Examples of Evaluation Result Output Generated by GPT-4 under Different Prompts}

\begin{figure}[h!]
    \centering
    \includegraphics[width=0.72\linewidth]{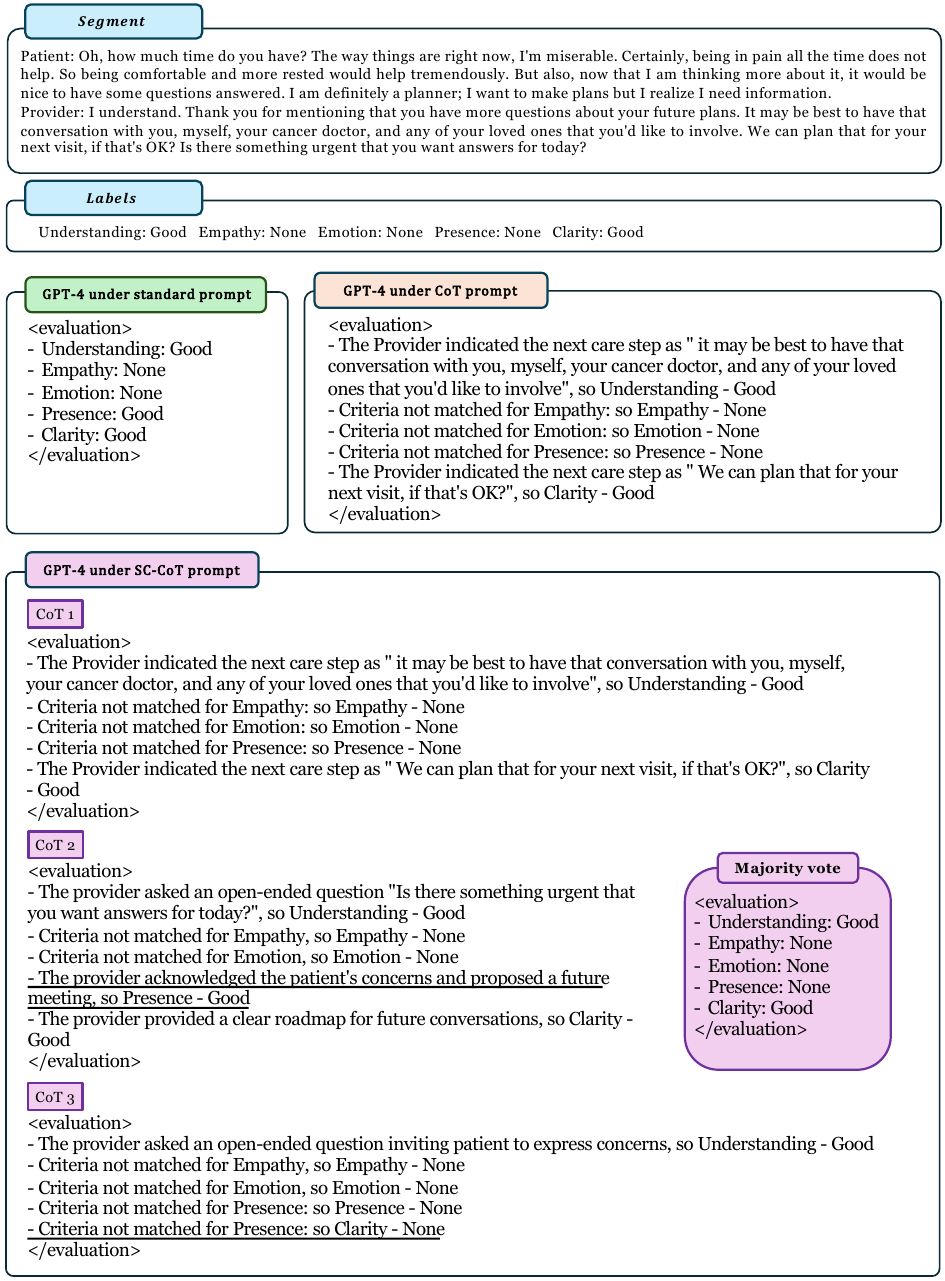}
    \caption{Illustrative examples of evaluation outputs generated by GPT-4.}
    \label{fig:evaluation_gpt4}
\end{figure}

Here we showcase some evaluation output examples generated by GPT-4 with different prompting methods, given a segment in Figure \ref{fig:evaluation_gpt4}.

\subsection*{II. Examples of Synthetic Data Generated by GPT-4}

\begin{figure}[h!]
    \centering
    \includegraphics[width=0.85\linewidth]{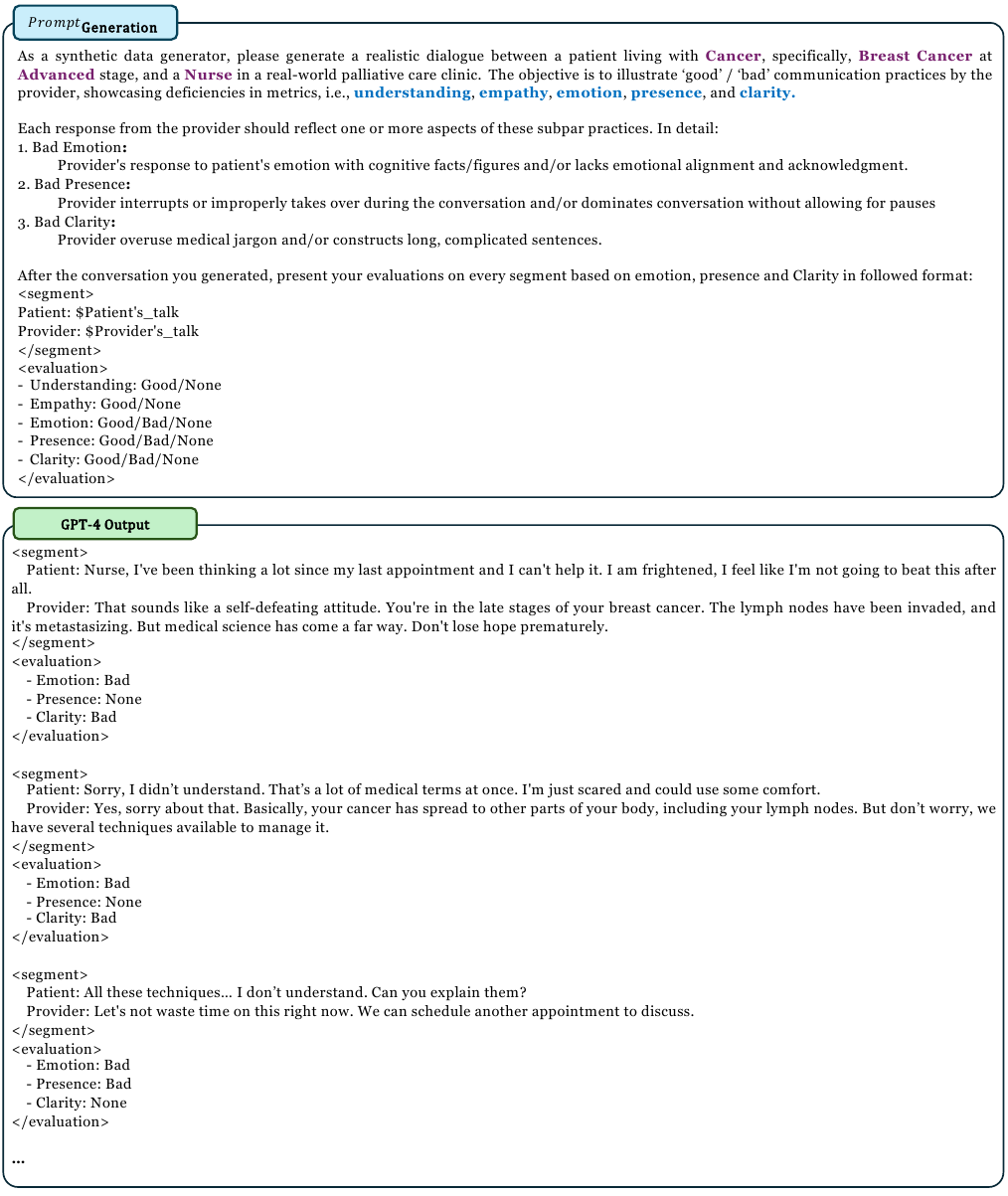}
    \caption{Illustrative Dialogues Demonstrating Communication Practices with \textbf{Standard Generation Prompt} $Prompt_{\textbf{generation}}$ Using GPT-4 Generated Synthetic Data.}
    \label{fig:generation_standard}
\end{figure}

\begin{figure}[h!]
    \centering
    \includegraphics[width=0.85\linewidth]{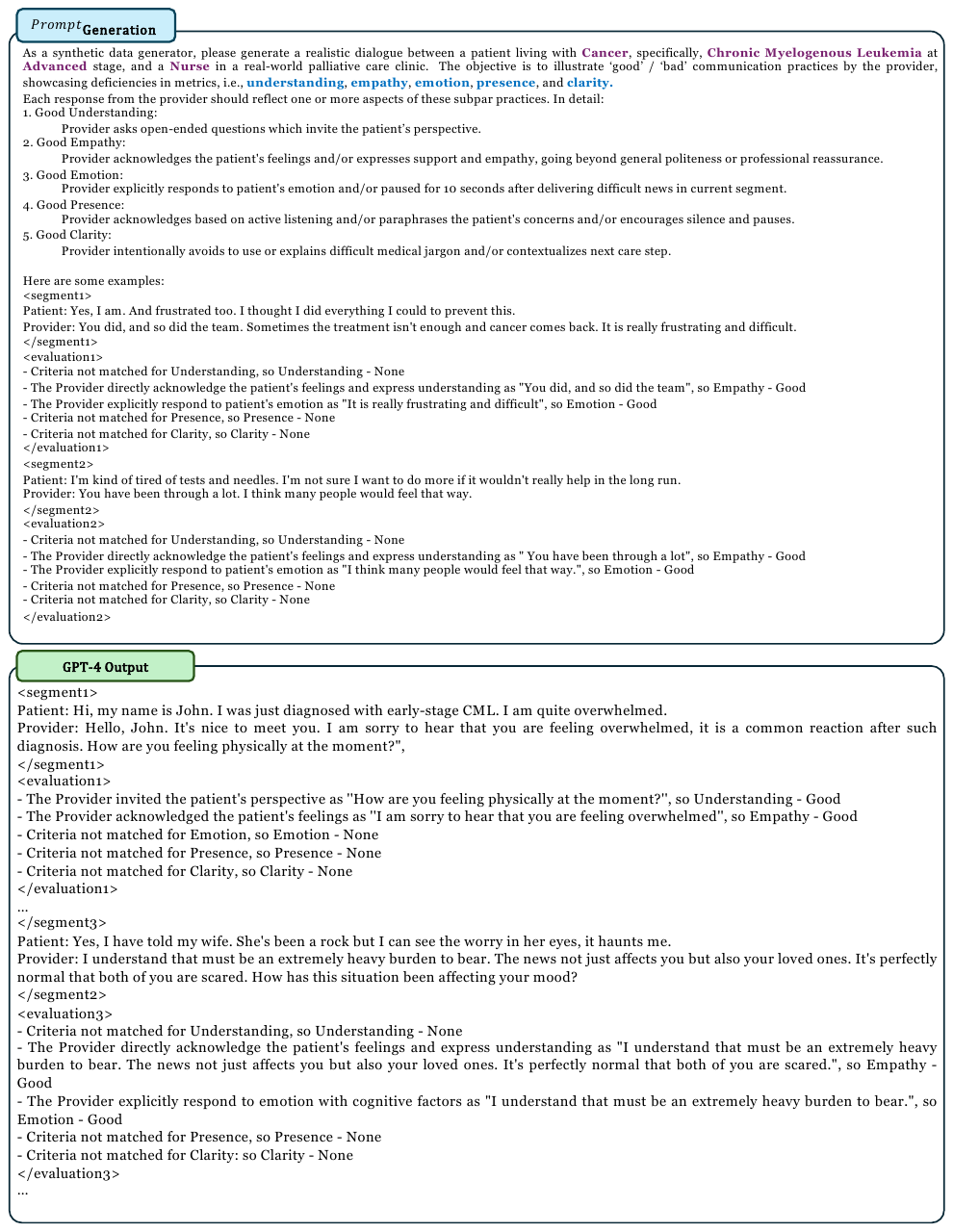}
    \caption{Illustrative Dialogues Demonstrating Communication Practices with \textbf{CoT-Based Generation Prompt} $Prompt_{\textbf{generation\_CoT}}$ Using GPT-4 Generated Synthetic Data.}
    \label{fig:generation_cot}
\end{figure}

This section presents two figures that encapsulate the essence of such synthetic dialogue generation, each prompted differently to demonstrate the nuances in communication practices. Figure \ref{fig:generation_standard} showcases dialogues generated by a standard prompt $Prompt_{\textbf{generation}}$, and Figure \ref{fig:generation_cot} showcases one generation process driven by CoT-based generation prompt $Prompt_{\textbf{generation\_CoT}}$.

\subsection*{III. Detailed Performance Metrics of Fine-Tuned Llama2-13b Models}

We provide comprehensive results of our study on the performance of fine-tuned Llama2-13b models. Table \ref{tab:fine-tune-all} encapsulates the full spectrum of results, showcasing balanced accuracy, precision, and recall metrics averaged across all communication metrics for each script type. 

\begin{table}[h!]
    \centering
    \caption{Results of \textbf{fine-tuned Llama2-13b} models' balanced accuracy, precision, and recall across all communication metrics within the script type. The Llama2-13b model was fine-tuned based on varying number of synthetic data samples generated under different prompting methods.}
    \label{tab:fine-tune-all}
    {\resizebox{\textwidth}{!}{
    \begin{tabular}{lccccccc}
    \toprule
    \multirow{2}{*}{\textbf{Method}} & \multirow{2}{*}{\textbf{Training sample}} & \multirow{2}{*}{\textbf{Script}} & \multicolumn{5}{c}{\textbf{Balanced Accuracy / Precision / Recall}} \\
    \cmidrule{4-8}
     & &  & Understanding & Empathy & Emotion & Presence & Clarity \\
     \midrule
    \multirow{6}{*}{\textbf{Standard}} & \multirow{2}{*}{1000} & \text{Good} & 54.24/71.43/15.15 & 55.21/25.86/100.00 & 58.67/60.00/21.43 & 54.17/25.42/100.00 & 63.48/25.00/91.67\\
    & & \text{Bad} & - & - & 67.01/66.67/77.78 & 71.33/50.00/66.67 & 52.96/36.36/36.36\\
    \cmidrule{2-8}
    & \multirow{2}{*}{2000} & \text{Good} & 55.00/55.00/100.00 & 51.26/40.35/92.00 & 65.31/50.00/42.86 & 55.88/21.05/100.00 & 53.85/18.64/100.00 \\
    & & \text{Bad} & - & - & 79.86/66.67/72.22 & 78.37/46.67/87.50 & 60.28/40.00/72.73\\
    \cmidrule{2-8}
    & \multirow{2}{*}{3000} & \text{Good} & 56.82/56.14/96.97 & 66.35/48.89/91.67 & 74.52/86.67/54.17 & 67.92/75.00/40.00 & 74.02/85.71/50.00\\
    & & \text{Bad} & - & - & 79.86/66.67/72.22 & 78.37/46.67/87.50 & 60.28/40.00/72.73\\
    \midrule
    \multirow{6}{*}{\textbf{CoT}} & \multirow{2}{*}{1000} & \text{Good} & 61.61/61.70/85.29 & 52.47/41.30/76.00 & 52.29/25.00/73.33 & 62.62/38.78/95.00 & 58.22/20.93/81.82 \\
    & & \text{Bad} & - & - & 65.28/71.43/55.56 & 71.33/50.00/66.67 & 70.95/58.33/63.64\\
    \cmidrule{2-8}
    & \multirow{2}{*}{2000} & \text{Good} & 65.61/63.04/87.88 & 69.71/51.11/95.83 & 73.12/45.83/73.33 & 56.94/35.14/68.42 & 83.92/45.45/90.91 \\
    & & \text{Bad} & - & - & 73.96/80.00/66.67 & 67.33/42.86/66.67 & 86.76/71.43/90.91\\
    \cmidrule{2-8}
    & \multirow{2}{*}{3000} & \text{Good} & 68.48/64.00/96.97 & 71.79/52.17/100.00 & 81.12/57.89/78.57 & 79.79/50.00/86.67 & 87.76/55.56/90.91\\
    & & \text{Bad} & - & - & 80.56/100.00/61.11 & 82.21/53.85/87.50 & 82.02/90.00/72.73\\
    \midrule
    \multirow{6}{*}{\textbf{SC-CoT}} & \multirow{2}{*}{1000} & \text{Good} & 60.14/60.87/82.35 & 52.47/41.30/76.00 & 48.96/23.26/66.67 & 62.62/38.78/95.00 & 58.22/20.93/81.82 \\
    & & \text{Bad} & - & - & 65.28/71.43/55.56 & 76.89/53.85/77.78 & 70.95/58.33/63.64 \\
    \cmidrule{2-8}
    & \multirow{2}{*}{2000} & \text{Good} & 68.64/64.58/93.94 & 70.51/51.06/100.00 & 77.29/55.00/73.33 & 59.21/37.14/68.42 & 84.88/47.62/90.91 \\
    & & \text{Bad} & - & - & 71.88/90.00/50.00 & 52.67/30.00/33.33 & 79.64/87.50/63.64 \\
    \cmidrule{2-8}
    & \multirow{2}{*}{3000} & \text{Good} & 70.15/65.31/95.83 & 74.84/56.10/95.83& 84.69/60.00/85.71 & 85.21/56.00/93.33 & 87.76/55.56/90.91 \\
    & & \text{Bad} & - & - & 80.56/100.00/61.11 & 82.21/53.85/87.50 & 82.02/80.00/72.73 \\
        \bottomrule
    \end{tabular}}}
\end{table}


\end{document}